\documentclass[mnsc]{informs3aa} 
\usepackage[final]{showkeys}
\usepackage{subfigure}
\usepackage{arydshln}
\usepackage{makecell}

\OneAndAHalfSpacedXI 

\usepackage{endnotes}

\usepackage{longtable}
\usepackage{tabularx}
\usepackage{booktabs}
\usepackage{etex}
\usepackage{multirow}
\usepackage{array}
\usepackage{bbm}

\usepackage[title]{appendix}

\newcommand{\vct}{\boldsymbol }

\newcommand{\op}{\mathrm{op}}

\renewcommand{\tilde}{\widetilde}
\renewcommand{\hat}{\widehat}
\renewcommand{\bar}{\overline}

\renewcommand{\hat}{\widehat}
\renewcommand{\tilde}{\widetilde}

\usepackage{amsbsy, amsfonts, amsgen, amsmath, amsopn, amssymb, amstext,
amsxtra, bezier, color, enumerate, graphicx, latexsym, verbatim,
pictexwd, supertabular, url, dsfont,leftidx, mathrsfs, appendix, setspace}
\usepackage{algorithm}
\usepackage{algorithmicx}
\usepackage{algpseudocode}
\makeatletter
\def\BState{\State\hskip-\ALG@thistlm}
\makeatother

\usepackage{pifont}

\usepackage{natbib}
 \bibpunct[, ]{(}{)}{,}{a}{}{,}%
 \def\bibfont{\small}%
\usepackage[colorlinks=true,bookmarks=false,urlcolor=blue, citecolor=blue,linkcolor=blue,bookmarksopen=false,draft=false]{hyperref}

\usepackage{parskip} 

\allowdisplaybreaks[4]

\TheoremsNumberedThrough     
\ECRepeatTheorems

\EquationsNumberedThrough    

\MANUSCRIPTNO{} 
\renewcommand{\hat}{\widehat}
\renewcommand{\tilde}{\widetilde}
\renewcommand{\bar}{\overline}

\newcommand{\alg}{{alg}}
\newcommand{\diam}{\mathrm{diam}}

\begin{document}


\RUNAUTHOR{Wang}

\RUNTITLE{On Adaptivity in Non-stationary Stochastic Optimizations}

\TITLE{On Adaptivity in Non-stationary Stochastic Optimization With Bandit Feedback}

\ARTICLEAUTHORS{%
\AUTHOR{Yining Wang}
\AFF{Naveen Jindal School of Management, University of Texas at Dallas, Richardson, TX 75070, USA}
} 

\ABSTRACT{In this paper we study the non-stationary stochastic optimization question with bandit feedback and dynamic regret measures.
The seminal work of \cite{besbes2015non} shows that, when aggregated function changes is known a priori, a simple re-starting algorithm attains the optimal dynamic regret.
In this work, we designed a stochastic optimization algorithm with fixed step sizes, which combined together with the multi-scale sampling framework of \cite{wei2021non}
achieves the optimal dynamic regret in non-stationary stochastic optimization \emph{without} requiring prior knowledge of function change budget, thereby closes a question that has been open for a while.
We also establish an additional result showing that any algorithm achieving good regret against \emph{stationary} benchmarks with high probability could be automatically converted to
an algorithm that achieves good regret against \emph{dynamic} benchmarks, which is applicable to a wide class of bandit convex optimization algorithms.

This version: \today
}



\date{}

\maketitle

\section{Introduction}

Sequential stochastic optimization with bandit feedback is a fundamental question in optimization and operations research,
with a wide range of applications in economics, computer science, engineering and revenue management research \citep{benveniste2012adaptive,lai2003stochastic}.
In its most basic form, an algorithm or optimizer interacts with an oracle sequentially over $T$ time periods. At the beginning of a time period $t$, the algorithm queries a solution 
$x_t\in\mathcal X$ with $\mathcal X$ being the optimization domain, and receives feedback $y_t\in\mathbb R$ satisfying 
$$
\mathbb E[y_t|x_t] = f(x_t),
$$
with $f:\mathcal X\to\mathbb R$ being an unknown objective function to be optimized.
The seminal work of \cite{besbes2015non} extends the sequential stochastic optimization problem to a \emph{non-stationary} setting, by allowing the functions to be optimized
over $T$ time periods to be slowly changing. More specifically, an unknown adversarial sequence of functions $f_1,\cdots,f_T:\mathcal X\to\mathbb R$ is selected by nature and the feedback $y_t$ received at time $t$ satisfies
$$
\mathbb E[y_t|x_t] = f_t(x_t).
$$
The objective is to design an algorithm that minimizes the expected \emph{dynamic regret}, defined as the expected sum of the differences between the solution points selected by the optimization algorithm
and the optimal solution for each function at each time period, or more specifically
\begin{equation}
\text{(dynamic regret)}\;\;\mathbb E\left[\sum_{t=1}^T f_t(x_t^*)-f_t(x_t)\right],\;\;\text{where}\;\;x_t^*=\arg\max_{x\in\mathcal X}f_t(x).
\label{eq:defn-dreg}
\end{equation}
Note that the dynamic regret is always non-negative by definition, and a smaller regret indicates an algorithm with better performance.

Because only partial information (noisy function evaluation at a single point) for each objective function $f_t$ is accessible, it is essential to limit the changes of unknown functions in order to ensure tractability.
In the work of \cite{besbes2015non}, it is assumed that 
\begin{equation}
\sum_{t=1}^{T-1}\|f_{t+1}-f_t\|_{\infty} \leq V_T\;\;\;\;\;\;\text{almost surely},
\label{eq:defn-Vt}
\end{equation}
where $V_T>0$ is a positive budget parameter upper bounding total function changes.
It is shown in \cite{besbes2015non} that, with Eq.~(\ref{eq:defn-dreg}) and additional smoothness and strong concavity assumptions, an algorithm based on frequent re-starts achieves dynamic regret upper bounded by
\begin{equation}
\mathbb E\left[\sum_{t=1}^T f_t(x_t^*)-f_t(x_t)\right]\leq C_1\times V_T^{1/3}T^{2/3}
\label{eq:dreg-scaling}
\end{equation}
for some constant $C_1<\infty$ independent of $V_T$ and $T$, which is information-theoretically optimal \citep{besbes2015non}.

\paragraph{\textbf{The adaptivity problem.}}
While the re-starting algorithm and its analysis in \citep{besbes2015non} are elegant and optimal, one major drawback is the requirement of prior knowledge of the changing budget $V_T$
in order to set up re-starting schedules, which is typically unknown in practical situations.
This drawback is also mentioned in the conclusion section of \citep{besbes2015non}, as a mis-specified parameter of $V_T$ would lead to degrading or even linear regret.
It has remained open whether \emph{adaptive} algorithms could be designed without prior knowledge of $V_T$ while still achieving the optimal $O(V_T^{1/3}T^{2/3})$ regret in Eq.~(\ref{eq:dreg-scaling}).

\paragraph{\textbf{Existing results.}}
There are several existing works making progress to this question, and partial solutions have been obtained with weaker regret guarantees or stronger modeling assumptions.
The work of \cite{zhang2018dynamic} analyzes an algorithm achieving the $O(V_T^{1/3}T^{2/3})$ regret without knowing the $V_T$ parameter under the \emph{online convex optimization (OCO)} 
setting, in which the entire objective function $f_t$ is revealed to the algorithm after each time period.
\cite{baby2022optimal} achieves a similar result with a more general algorithmic framework.
Such a setting is difficult to justify as in most application scenarios (e.g., dynamic pricing) only noisy evaluation at $f_t(x_t)$ is available.
The work of \cite{cheung2022hedging} uses a multiplicative weight update procedure to ``hedge'' several algorithmic threads to achieve adaptivity in $V_T$.
Unfortunately, the approach incurs an additional $O(T^{3/4})$ in the regret bound which is sub-optimal for general problems.
The work of \cite{wei2021non} proposed a general multi-scale sampling framework that achieves near-optimal regret guarantee for several bandit problems
(multi-armed bandit, linear bandit, reinforcement learning, etc.) without knowing $V_T$,
provided that the sub-routine used satisfies certain ``UCB'' properties.
It is remarked in the conclusion section of \cite{wei2021non} that they are not aware of near-optimal bandit convex optimization algorithms satisfying such UCB properties,
rendering the framework not directly applicable to the non-stationary stochastic optimization problem.

\paragraph{\textbf{Our contributions.}}
In this paper, we build upon the framework of \cite{wei2021non} to achieve the $\tilde O(V_T^{2/3}T^{1/3})$ regret in bandit optimization without knowing the changing budget $V_T$ a priori.
At the core of the algorithm is a carefully designed estimating gradient descent method with \emph{fixed} step sizes and batch gradient estimation, which we prove satisfies certain UCB properties proposed in \citep{wei2021non} in weakly non-stationary environments. Our designed algorithm is different from standard approach using \emph{diminishing} step sizes and single-point gradient estimation \citep{flaxman2005online,agarwal2010optimal,besbes2015non}, 
which does not have good regret upper bounds with high probability.

We also derive an interesting result (Sec.~\ref{sec:connection}) showing that any bandit optimization algorithm that enjoys good \emph{stationary regret} (comparing against a stationary benchmark)
upper bound with high probability and is ``anytime'' (satisfying regret upper bounds continuously without knowing time horizon a priori) can be automatically converted to an algorithm
satisfying UCB properties \citep{wei2021non}, and therefore could be used for dynamic regret in non-stationary environments.
While no existing bandit convex optimization algorithm explicitly satisfies such conditions, the connection between stationary regret and dynamic regret is interesting
and could inspire the design of adaptively non-stationary algorithms for other problems.

\subsection{Assumptions and notations}\label{subsec:assumptions}

Apart from the non-stationary constraint in Eq.~(\ref{eq:dreg-scaling}), we impose the following additional assumptions on the functions $f_1,\cdots,f_T$:
\begin{enumerate}
\item \emph{(Bounded rewards and domain).} It holds that $|f_t(x)|\leq 1$ and $|y_t|\leq 1$ almost surely for every $t$ and $x\in\mathcal X$; furthermore, the domain $\mathcal X\subseteq\mathbb R^d$
is compact with $\diam(\mathcal X)=\max_{x,x'\in\mathcal X}\|x-x'\|_2\leq B_X<\infty$;
\item \emph{(Smoothness).} Each $f_t$ is twice continuously differentiable on $\mathcal X$ and satisfies for every $x\in\mathcal X$ that $\max\{\|\nabla f_t(x)\|_2, \|\nabla^2 f_t(x)\|_{\op}\}\leq L$
for some constant $L<\infty$;
\item \emph{(Strong concavity).} There exists a constant $\sigma>0$ such that for each $f_t$ and $x\in\mathcal X$, $-\nabla^2 f_t(x)\succeq \sigma I$;
\item \emph{(Interior maximizer).} There exists a constant $c_0\in(0,1)$ such that for each $f_t$, $x_t^*=\arg\max_{x\in\mathcal X}f_t(x)$ satisfies
$\|x_t^*-z\|_2\geq c_0$ for all $z\in\partial \mathcal X$.
\end{enumerate}

The above assumptions are conventional in the bandit convex optimization literature; see e.g.~the work of \cite{besbes2015non}. 

We further define some notations used throughout this paper. For $z\in\mathbb R^d$ and $r>0$, let $\mathbb B_2(z,r)=\{x\in\mathbb R^d: \|x-z\|_2\leq r\}$ denote the $\ell_2$-ball
centered at $z$ with radius $r$.
For a convex compact set $\mathcal Z\subseteq\mathbb R^d$ and a real number $r>0$, let $\mathcal Z^o(r)=\{x: \mathbb B_2(x,r)\subseteq\mathcal Z\}$ be the interior of $\mathcal Z$
that is at least $r$ distance away from its boundary.
Assumption 4 can then be written as $x_t^*\in \mathcal X^o(c_0)$ for every $t$.
Finally, for a convex compact set $\mathcal Z\subseteq\mathbb R^d$, let $P_{\mathcal Z}(\cdot) = \arg\min_{z\in\mathcal Z}\|z-\cdot\|_2$ be the projection operator onto $\mathcal Z$.
For a sequence $\{a_n\}_{n\geq 1}$, we write $a_n=O(f(n))$ if $\limsup_{n\to\infty}|a_n|/|f(n)|<\infty$, and $a_n=\tilde O(f(n))$ if there exists $\gamma>0$ such that $\limsup_{n\to\infty}|a_n|/|f(n)\ln^\gamma n|<\infty$.

\subsection{Other related works}

Stochastic zeroth-order optimization and its bandit convex optimization extension have attracted an enormous amount of interest in the computer science and operations research literature.
The key challenges of these types of questions are the lack of accurate or unbiased first-order information, rendering most first-order methods not directly applicable.
Existing approaches when the underlying objective function does not change include cone-cutting type methods \citep{agarwal2013stochastic,nemirovskij1983problem},
ellipsoidal methods \citep{lattimore2021improved}, and in some special cases first-order methods with estimated gradients \citep{agarwal2010optimal,wang2018stochastic}.
When the underlying functions change, kernel based methods and existential arguments are employed \citep{bubeck2021kernel,lattimore2020improved}.
In the computer science literature it is usually the stationary benchmark that is studied for fully adversarial environments, while dynamic regret with function change budgets is much less investigated.

Another direction of research that is closely related to the bandit convex optimization question is \emph{online convex optimization}, in which the algorithm observes the entire function $f_t$ after it produces an estimate $x_t\in\mathcal X$
at time $t$. Clearly, the online convex optimization question is easier compared with optimization problem with bandit feedback studied in this paper, because the full information of the objective function $f_t$ is available after time $t$.
This also means that some more aggressive non-stationary regret measures, such as the \emph{adaptive regret} \emph{hazan2009efficient} which measures the regret of an adaptive algorithm on any arbitrary intervals.
Many existing works have studied dynamic and adaptive regret for such full-information problems, including settings in which the function changing budget is unknown.
See for example the works of \cite{baby2022optimal,zhao2022non,zhao2020dynamic,zhang2020minimizing,lu2022non}, and many more.

\section{Non-stationary bandit optimization without prior knowledge}

The work of \cite{wei2021non} proposed a general framework for non-stationary bandit optimization without prior knowledge of changing budget.
In this section we summarize the proposed framework and also theoretical conditions/properties for regret analysis.
We clarify that all definitions and results in this section are from the work of \cite{wei2021non} and therefore known,
and our summary here is merely for the completeness of this paper.

For each $t\in\{1,2,\cdots,T-1\}$, let $\Delta(t) := \|f_{t+1}-f_t\|_{\infty}$ be the measure of function change from time $t$ to time $t+1$.
For an interval $I=[s,t]$, let $\Delta_I := \sum_{\tau=s}^{t-1}\Delta(\tau)$.
Clearly, $\Delta_{[1,T]} \leq V_T$ thanks to Eq.~(\ref{eq:defn-Vt}).
The algorithm design and analysis of \cite{wei2021non} were built upon the concept of ``weakly non-stationary'' algorithms, which is summarized in the following condition:

\begin{definition}
Let $\rho(\cdot)$ be a decreasing function so that $C(t)=t\rho(t)$ is increasing, and $\lambda>0$ be a scaling parameter.
A weakly non-stationary bandit optimization algorithm $\mathcal A$ produces an estimate $\bar r_t\in\mathbb R$ at the end of each time period $t$, such that the following holds:
for any $t\in[T]$ such that $\Delta_{[1,t]}\leq\rho(t)/\lambda$, it holds with probability $1-\tilde O(T^{-2})$ that
\begin{enumerate}
\item $\bar r_t\geq \min_{\tau\leq t}f_\tau^* - \lambda\Delta_{[1,t]}$, where $f_\tau^*=\max_{x\in\mathcal X}f_\tau(x)$;
\item $\frac{1}{t}\sum_{\tau=1}^t (\bar r_t-y_t) \leq \rho(t) + \lambda\Delta_{[1,t]}$.
\end{enumerate}
\label{defn:weakly-nonstationary}
\end{definition}

Intuitively, a weakly non-stationary algorithm $\mathcal A$ produces an upper-confidence estimate $\bar r_t$ at the end of each time period which both upper bounds the optimal reward 
and is close to the actual rewards collected on the path.
With such an algorithm $\mathcal A$, the multi-scale sampling framework of \cite{wei2021non} is summarized in Algorithm \ref{alg:sampling}.

\begin{algorithm}[t]
\caption{A multi-scale sampling framework}
\label{alg:sampling}
\begin{algorithmic}[1]
\State \textbf{Input}: time horizon $T$, the $\rho(\cdot)$ function in Definition \ref{defn:weakly-nonstationary}, weakly non-stationary algorithm $\mathcal A$.
\State Initialize: $\hat\rho(\cdot)=6(\log_2 T+1)\rho(\cdot)$, $t\gets 1$;
\For{$n=0,1,2,\cdots$}\label{line:outer-for}
	\State Set $t_n\gets t$. For each $m=n,n-1,\cdots,0$ and $\tau=t_n+z\cdot 2^m$, $z\in\{0,1,\cdots,2^{n-m}-1\}$, 
schedule an order-$m$ thread $\alg$ as a copy of the weakly non-stationary algorithm $\mathcal A$ starting at $\alg.s=\tau$ ending at $\alg.e=\tau+2^m$ with probability $\rho(2^n)/\rho(2^m)$;
	\While{$t<t_n+2^n$}
		\State Run the lowest order thread schedued covering $t$ and obtain $y_t,\bar r_t$;
		\State Set $U_t=\min_{\tau\in[t_n,t]}\bar r_t$;
		\State Perform Test 1 and Test 2 and increment $t\gets t+1$;
		\State If either test returns fail, restart from Line \ref{line:outer-for} with $n=0$;
	\EndWhile
\EndFor
\end{algorithmic}
\textbf{Test 1}: if $t=\alg.e$ for some scheduled order-$m$ thread $\alg$ and $\frac{1}{2^m}\sum_{\tau=alg.s}^{alg.e}y_\tau\geq U_t+9\hat\rho(2^m)$, return fail;\\
\textbf{Test 2}: of $\frac{1}{t-t_n+1}\sum_{\tau=t_n}^t(\bar r_\tau-y_\tau)\geq 3\hat\rho(t-t_n+1)$, return fail.
\end{algorithm}

The following regret upper bound is established in \cite[Theorem 2]{wei2021non}.
\begin{lemma}
Suppose $C(t)=t\rho(t)$ satisfies $C(t)=c_1\sqrt{t}+c_2$ for some constants $c_1,c_2>0$. Then the dynamic regret of Algorithm \ref{alg:sampling} can be upper bounded by
\begin{align*}
\mathbb E\left[\sum_{t=1}^T f_t(x_t^*)-f_t(x_t)\right]
\leq \widetilde O\left(C_1V_T^{1/3}T^{2/3} + C_2\sqrt{T}\right),
\end{align*}

where $C_1=(c_1^{2/3}+c_2c_1^{-4/3})\lambda^{2/3}$ and $C_2=c_1+c_2/c_1$ are polynomial functions of $c_1,c_2,\lambda$, and in $\widetilde O(\cdot)$ notation we omit all poly-logarithmic factors.
\label{lem:multi-scale}
\end{lemma}

\begin{remark}
When $C(t)$ takes on forms other than $c_1\sqrt{t}+c_2$ similar regret upper bounds could be derived in \cite{wei2021non}. However, we focus exclusively
on the $C(t)=c_1\sqrt{t}+c_2$ form in this paper because that is the form corresponding to bandit convex optimization problems.
\end{remark}

\section{Main results}

In this section we present the main results of this paper. We first present a revised estimating gradient descent algorithm
that differs from existing approaches in terms of both gradient estimation procedures and step size strategies.
We then analyze the proposed algorithm in weakly non-stationary scenarios, and show that the algorithm satisfies the properties listed in Definition \ref{defn:weakly-nonstationary}.
Tight regret upper bounds are then implied.

\subsection{A revised estimating gradient ascent algorithm}

In Algorithm \ref{alg:inexact-prox} we present a revised estimating gradient ascent algorithm that works with bandit feedback.
At a higher level, the algorithm is divided into batches $s=0,1,2,\cdots$ of geometrically increasing lengths.
Within each batch, the algorithm repeatedly explores each dimension for $2n_s$ periods at a $\delta_s$ step size to estimate the gradient $\nabla f(z_s)$, 
with $n_s$ and $\delta_s$ being carefully selected so that the estimation errors of $\nabla f(z_s)$ also decrease with $s$.
The algorithm then carries out projected gradient ascent algorithm (Line \ref{line:pga}) with \emph{fixed} step size $\eta_0$,
which converges to the optimal solution of the objective function at a fast pace.

\begin{algorithm}[t]
\caption{A weakly non-stationary algorithm for bandit optimization}
\label{alg:inexact-prox}
\begin{algorithmic}[1]
\State \textbf{Input}: parameters $\eta_0, \kappa_0>0$, $\gamma\in(0,1)$, $c_0>0$, $N_0\in\mathbb N$.
\State \textbf{Initialize}: An arbitrary $z_0\in\mathcal X^o(c_0)$, $t_0= 1$, $r= 0$;
\For{$s=0,1,2,\cdots$}
	\For{$j=1,2,\cdots,d$}
		\State Let $n_s = \lceil(1-\gamma)^{-4s}N_0\rceil$, $\delta_s= n_s^{-1/4}$, $u_+= u_-= 0$;
		\For{$t=t_0,t_0+1,\cdots,t_0+2n_s-1$}
			\If{$t-t_0$ is an even number}
				\State Take action $x_t=z_s+\delta_s e_j$ and observe feedback $y_t$;
				\State Increment $u_+\gets u_++y_t$, $r\gets r+y_t$ and output $\bar r_t \gets (r/t) + 2\kappa_0/\sqrt{t}$;
			\Else
				\State Take action $x_t=z_s-\delta_s e_j$ and observe feedback $y_t$;
				\State Increment $u_-\gets u_-+y_t$, $r\gets r+y_t$ and output $\bar r_t \gets (r/t) + 2\kappa_0/\sqrt{t}$;
			\EndIf
		\EndFor
		\State Let $\hat g_s(j) = (u_+ - u_-)/(2\delta_sn_s)$;
	\EndFor
	\State Update $z_{s+1} = P_{\mathcal X^o(c_0)}(z_s + \eta_0 \hat g_s)$, where $\hat g_s=[\hat g_s(1),\cdots,\hat g_s(d)]\in\mathbb R^d$.\label{line:pga}
\EndFor
\end{algorithmic}
\end{algorithm}

\subsection{Analysis in weakly non-stationary environments}

The main purpose of this section is to analyze the performance of Algorithm \ref{alg:inexact-prox} in weakly non-stationary environments,
and show that both properties in the important Definition \ref{defn:weakly-nonstationary} are satisfied.

\begin{lemma}
Suppose Algorithm \ref{alg:inexact-prox} is executed with $\eta_0=1/L$, $\kappa_0= \sqrt{\ln(dT)} + L+\frac{2d\ln T}{1-\gamma} + \frac{16B_X^2d^{3/2}\ln^4(dT)}{c_0(1-\gamma)^3} + \frac{2L^2d^{3/2}\ln^4(dT)}{c_0(1-\gamma)^7}$, $\gamma=\sigma/L$ and $N_0=\lceil c_0^{-2}\rceil + 1$. Then the estimates $\{\bar r_t\}_{t=1}^T$ output by Algorithm \ref{alg:inexact-prox}
satisfy Definition \ref{defn:weakly-nonstationary} with $\rho(t)=6\kappa_0/\sqrt{t}$ and $\lambda=6\kappa_0$.
\label{lem:inexact-prox}
\end{lemma}
\begin{remark}
In practice, the problem parameters $L$, $\sigma$ and $c_0$ might be unknown.
In such cases, the values of these parameters could be set as slowing increasing or decreasing functions of $T$, such as $L,B_X=\ln T$ and $\sigma,c_0=1/\ln T$.
Note that such a strategy cannot be applied to the unknown changing budget $V_T$ in general, because $V_T$ involves $T$ sequential functions and is typically scales as a polynomial function of $T$ which is different from other problem parameters $L,\sigma,c_0,B_X$ applicable to single functions, meaning that they remain fixed constants in most cases.
\end{remark}

\begin{proof}{Proof of Lemma \ref{lem:inexact-prox}.}
First, note that with $N_0\geq c_0^{-2}$ it holds that $x_t\in\mathcal X$ almost surely for all $t$, because $z_s\in\mathcal X^o(c_0)$. This shows that all solutions provided by the algorithm
are feasible with probability one.

Let $\bar t$ be the target number of time periods in Definition \ref{defn:weakly-nonstationary}.
Throughout this proof, we abbreviate $\Delta := \Delta_{[1,\bar t]}$ to simplify notations.
For each epoch $s$ and coordinate $j\in[d]$, let $\varepsilon_s(j) := g_s(j)-\hat g_s(j)$ be the estimation error of gradient, where $g_s =(g_s(1),\cdots,g_s(d))= \nabla f_1(z_s)$ is the gradient of the objective function
at the first time period. Let also $\mathcal T_s^{\pm}(j)=\mathcal T_s^+(j)\cup\mathcal T_s^-(j)$ be the set of all time periods devoted to solution $x_t=z_s\pm\delta_s e_j$, which satisfies $|\mathcal T_s^{\pm}(j)|=|\mathcal T_s^+(j)|+|\mathcal T_s^-(j)|=2n_s=\lceil (1-\gamma)^{-4s} N_0\rceil$.
By Hoeffding-Azuma's inequality \citep{hoeffding1994probability,azuma1967weighted}, it holds with probability $1-\tilde O(d^{-1}T^{-2})$ that
\begin{equation}
\left|\frac{u_+}{n_s}-\frac{1}{n_s}\sum_{\tau\in\mathcal T_s^+(j)}f_\tau(z_s+\delta_s e_j)\right|\leq 2\sqrt{\frac{\ln(dT)}{n_s}}.
\label{eq:proof-inexact-1}
\end{equation}
Note that for every $\tau$, $\|f_\tau-f_1\|_{\infty}\leq \Delta_{[1,\tau]}\leq\Delta$ by definition. Subsequently, Eq.~(\ref{eq:proof-inexact-1}) implies
\begin{equation}
\left|\frac{u_+}{n_s} - f_1(z_s+\delta_s e_j)\right|\leq \Delta + 2\sqrt{\frac{\ln(dT)}{n_s}}.
\label{eq:proof-inexact-2}
\end{equation}
By a symmetric argument, it holds with probability $1-\tilde O(d^{-1}T^{-2})$ that 
\begin{equation}
\left|\frac{u_-}{n_s} - f_1(z_s-\delta_s e_j)\right|\leq \Delta +2 \sqrt{\frac{\ln(dT)}{n_s}}.
\label{eq:proof-inexact-3}
\end{equation}
By Assumption 2 and Taylor expansion, we have that
\begin{equation}
\big|f_1(z_s+\delta_s e_j)-f_1(z_s-\delta_s e_j)- 2\delta_s\partial_j f_1(z_s)\big|\leq L\delta_s^2.
\label{eq:proof-inexact-4}
\end{equation}
Combining Eqs.~(\ref{eq:proof-inexact-2},\ref{eq:proof-inexact-3},\ref{eq:proof-inexact-4}) and noting that $g_s(j) = \partial_j f_1(z_s)$, we obtain
\begin{equation}
\big|\hat g_s(j)-g_s(j)\big| \leq \frac{1}{\delta_s}\left(\Delta + 2\sqrt{\frac{\ln(dT)}{n_s}}\right) + \frac{L}{2}\delta_s\leq \Delta n_s^{1/4} +2 n_s^{-1/4}\sqrt{\ln(dT)} + Ln_s^{-1/4},
\label{eq:proof-inexact-5}
\end{equation}
where the last inequality holds because $\delta_s=n_s^{-1/4}$ as specified in Algorithm \ref{alg:inexact-prox}.
Let $\varepsilon_s := (\varepsilon_s(1),\cdots,\varepsilon_s(d))\in\mathbb R^d$. By a union bound, we have with probability $1-\tilde O(T^{-2})$ that
\begin{align}
\|\varepsilon_s\|_2 &= \|\hat g_s-\nabla f_1(z_s)\|_2 \leq (\Delta n_s^{1/4} + 2n_s^{-1/4}\sqrt{\ln(dT)} + Ln_s^{-1/4})\sqrt{d}\nonumber\\
&\leq (1-\gamma)^{-s}N_0^{1/4}\sqrt{d}\Delta +3 (1-\gamma)^sL\sqrt{d\ln(dT)}.
\label{eq:proof-inexact-6}
\end{align}

Recall the definition that $\gamma = \sigma/L > 0$, where $\sigma,L$ are parameters in the assumptions imposed (see Sec.~\ref{subsec:assumptions}).
Let also $x_1^*=\arg\max_{x\in\mathcal X}f_1(x)$ be the maximizer of $f_1$, which satisfies $x_1^*\in\mathcal X^o(c_0)$ thanks to Assumption 4.
Note that $\|z_0-x_1^*\|_2\leq \diam(\mathcal X)\leq B_X$, where $B_X$ is defined in Assumption 1. Citing \citep[Proposition 1]{schmidt2011convergence}, it holds with probability 1 that
\begin{align}
\|z_s-x_1^*\|_2 &\leq (1-\gamma)^sB_X + \sum_{j=1}^s(1-\gamma)^{s-j}\frac{\|\varepsilon_j\|_2}{L}\nonumber\\
&\leq (1-\gamma)^s B_X + \sum_{j=1}^s (1-\gamma)^{s-2j}N_0^{1/4}\sqrt{d}\Delta + \sum_{j=1}^s 3(1-\gamma)^s \sqrt{d\ln(dT)}\label{eq:proof-inexact-7}\\
&\leq (1-\gamma)^s B_X + s(1-\gamma)^{-s}N_0^{1/4}\sqrt{d}\Delta +3 s(1-\gamma)^s \sqrt{d\ln(dT)}\nonumber\\
&\leq \frac{B_XN_0^{1/4}}{n_s^{1/4}} + sn_s^{1/4}\sqrt{d}\Delta + \frac{3s N_0^{1/4}\sqrt{d\ln(dT)}}{n_s^{1/4}}\label{eq:proof-inexact-8}\\
&\leq \frac{(B_X+3s+s\sqrt{n_s}\Delta)N_0^{1/4}\sqrt{d\ln(NT)}}{n_s^{1/4}}.\label{eq:proof-inexact-9}
\end{align}
Here, Eq.~(\ref{eq:proof-inexact-7}) holds by incorporating Eq.~(\ref{eq:proof-inexact-6}); 
Eq.~(\ref{eq:proof-inexact-8}) holds because $n_s=\lceil (1-\gamma)^{-4s}N_0\rceil$.
Because $x_1^*$ is an interior maximizer and $f_1$ is strongly concave, we have $\nabla f_1(x_1^*)=0$ and subsequently
\begin{align}
f_1(x_1^*)-f_1(z_s) &\leq \frac{\sigma}{2}\|x_1^*-z_s\|_2^2 \leq \frac{(2B_X^2+14s^2+2s^2 n_s\Delta^2)\sqrt{N_0}d\ln(NT)}{\sqrt{n_s}}.
\label{eq:proof-inexact-9half}
\end{align}

Let $\bar s$ be the epoch containing the target time period $\bar t$. First note that
\begin{align}
&\left|\sum_{t=1}^{\bar t} (y_t - f_1(x_1^*))\right| \leq
\left|\sum_{s=1}^{\bar s}\sum_{j=1}^d\sum_{t\in\mathcal T_s^{\pm}(j),t\leq\bar t}(y_t-f_1(x_1^*))\right|\nonumber\\
&\leq \underbrace{\left|\sum_{t=1}^{\bar t}(y_t-f_t(x_t))\right|}_{\mathcal A} + \underbrace{\sum_{t=1}^{\bar t}\big|f_t(x_t)-f_1(x_t)\big|}_{\mathcal B} + \underbrace{\sum_{s=0}^{\bar s}\min\{2dn_s,\bar t\}\times \big|f_1(z_{s})-f_1(x_1^*)\big|}_{\mathcal C}\nonumber\\
&+ \underbrace{\sum_{s=0}^{\bar s}\sum_{j=1}^d \sum_{t\in\mathcal T_s^+(j),t\leq\bar t}\big|f_1(z_s+\delta_se_j)+\vct 1\{t+1\leq\bar t\}f_1(z_s-\delta_s e_j) - (1+\vct 1\{t+1\leq\bar t\})f_1(z_s)\big|}_{\mathcal D}.\label{eq:proof-inexact-7}
\end{align}

We next proceed to upper bound the four terms of $\mathcal A,\mathcal B,\mathcal C,\mathcal D$ in Eq.~(\ref{eq:proof-inexact-7}) seperately.
For term $\mathcal A$, invoking the Hoeffding-Azuma inequality we have with probability $1-\tilde O(T^{-2})$ that
\begin{equation}
\mathcal A = \left|\sum_{t=1}^{\bar t}(y_t-f_t(x_t))\right| \leq \sqrt{\bar t\ln(dT)}.
\label{eq:proof-inexact-10}
\end{equation}
For term $\mathcal B$, note that $\|f_t-f_1\|_{\infty}\leq \Delta_{[1,t]}\leq \Delta_{[1,\bar t]}=\Delta$. Therefore,
\begin{equation}
\mathcal B = \sum_{t=1}^{\bar t}\big|f_t(x_t)-f_1(x_t)\big| \leq \bar t\Delta.
\label{eq:proof-inexact-11}
\end{equation}
For term $\mathcal C$, invoke Eq.~(\ref{eq:proof-inexact-9half}) and note that $\bar s\leq \ln T/(-4\ln(1-\gamma))\leq \ln T/(1-\gamma)$ because each epoch contains at least $(1-\gamma)^{-4}$ time periods,
and there are $T$ time periods in total. Subsequently,
\begin{align}
\mathcal C &\leq \sum_{s=0}^{\bar s}\min\{2dn_s,\bar t\}\times \frac{(2B_X^2+14s^2+2s^2 n_s\Delta^2)\sqrt{N_0}d\ln(NT)}{\sqrt{n_s}}\nonumber\\
&\leq \sum_{s=0}^{\bar s} (2B_X^2+14s^2+2s^2(1-\gamma)^{-4}\bar t\Delta^2)\sqrt{\tilde n_s}\times d^{3/2}\sqrt{N_0}\ln(dT)\label{eq:proof-inexact-12}\\
&\leq (2B_X^2+14\bar s^2+2\bar s^2(1-\gamma)^{-4}\bar t\Delta^2)\bar s\sqrt{\bar t}\times d^{3/2}\sqrt{N_0}\ln(dT)\nonumber\\
&\leq (16B_X^2 + 2(1-\gamma)^{-4}\bar t\Delta^2)\times d^{3/2}\bar s^3\sqrt{N_0\bar t}\ln(dT)\nonumber\\
&\leq (16B_X^2 + 2(1-\gamma)^{-4}\bar t\Delta^2)\times d^{3/2}(1-\gamma)^{-3}c_0^{-1}\sqrt{\bar t}\ln^4(dT).\label{eq:proof-inexact-13}
\end{align}
Here, in Eq.~(\ref{eq:proof-inexact-12}) we define $\tilde n_s := \min\{2dn_s,\bar t\}$ and replace $2s^2n_s\Delta^2$ with $2s^2\bar t(1-\gamma)^{-4}\Delta^2$ because $n_s\leq \bar t$ for $s<\bar s$,
and $n_{\bar s}\leq (1-\gamma)^{-4}n_{\bar s-1}\leq (1-\gamma)^{-4}\bar t$;
Eq.~(\ref{eq:proof-inexact-13}) holds by plugging in the upper bounds on $\bar s$ and $N_0$.
Finally, for term $\mathcal D$, note that for each $z_s$, $\delta_s$ and $j\in[d]$, we have 
\begin{align*}
\big|f_1(z_s+\delta_s)-f_1(z_s)-\delta_s \partial_j f_1(z_s)\big| &\leq L^2\delta_s^2/2 = L^2n_s^{-1/2}/2;\\
\big|f_1(z_s-\delta_s)-f_1(z_s)+\delta_s \partial_j f_1(z_s)\big| &\leq L^2\delta_s^2/2 = L^2n_s^{-1/2}/2.
\end{align*}
Subsequently, 
\begin{align}
\mathcal D &\leq L\delta_{\bar s} + \sum_{s=0}^{\bar s}2dn_s\times L^2n_s^{-1/2}\leq L + 2\bar sd\sqrt{\bar t}\leq L + 2(1-\gamma)^{-1}d\sqrt{\bar t}\ln T.
\label{eq:proof-inexact-14}
\end{align}

Combine Eqs.~(\ref{eq:proof-inexact-10},\ref{eq:proof-inexact-11},\ref{eq:proof-inexact-13},\ref{eq:proof-inexact-14}). Dividing both sides of Eq.~(\ref{eq:proof-inexact-7}) by $\bar t$, we obtain
with probability $1-\tilde O(T^{-2})$ that
\begin{align}
\left|\frac{1}{\bar t}\sum_{t=1}^{\bar t} (y_t - f_1(x_1^*))\right|
&\leq \sqrt{\frac{\ln(dT)}{\bar t}} + \Delta + \frac{L}{\bar t} + \frac{2d\ln T}{(1-\gamma)\sqrt{\bar t}} + \frac{16B_X^2d^{3/2}\ln^4(dT)}{c_0(1-\gamma)^3\sqrt{\bar t}}\nonumber\\
&\;\;\;\; + \frac{2L^2d^{3/2}\ln^4(dT)}{c_0(1-\gamma)^{7}\sqrt{\bar t}}\times \bar t\Delta^2.\label{eq:proof-inexact-15}
\end{align}

With the definitions that
\begin{align*}
\kappa_0 &= \sqrt{\ln(dT)} + L+\frac{2d\ln T}{1-\gamma} + \frac{16B_X^2d^{3/2}\ln^4(dT)}{c_0(1-\gamma)^3} + \frac{2L^2d^{3/2}\ln^4(dT)}{c_0(1-\gamma)^7};\\
\rho(t) &= 6\kappa_0/\sqrt{\bar t};\;\;\;\;\;\;\lambda = 6\kappa_0,
\end{align*}
the inequality (\ref{eq:proof-inexact-15}), which holds with probability $1-\tilde O(T^{-2})$, can be simplified to
\begin{equation}
\left|\frac{1}{\bar t}\sum_{t=1}^{\bar t} (y_t - f_1(x_1^*))\right| \leq \frac{(1+\kappa_0)}{\sqrt{\bar t}}\leq \frac{2\kappa_0}{\sqrt{\bar t}}.
\label{eq:proof-inexact-15-half}
\end{equation}
Note that the statistics $\{\bar r_t\}$ returned by Algorithm \ref{alg:inexact-prox} can be written as $\bar r_t = (\sum_{\tau\leq t}y_\tau)/t + 2\kappa_0/\sqrt{t}$.
The above inequality immediately implies the first property in Definition \ref{defn:weakly-nonstationary}. 
To prove the second property in Definition \ref{defn:weakly-nonstationary} for $\bar t$, we use the following derivation which holds with probability $1-\tilde O(T^{-1})$:
\begin{align}
\left|\frac{1}{\bar t}\sum_{t=1}^{\bar t}\bar r_t-y_t\right|
&\leq \left|\frac{1}{\bar t}\sum_{t=1}^{\bar t}\left[\left(\frac{1}{t}\sum_{\tau=1}^t y_\tau\right) - f_t(x_t)\right]\right| + \frac{1}{\bar t}\sum_{t=1}^{\bar t}\frac{2\kappa_0}{\sqrt{t}}\label{eq:proof-inexact-16}\\
&\leq \left|\frac{1}{\bar t}\sum_{t=1}^{\bar t} f_1(x_1^*) - f_t(x_t)\right| + \frac{1}{\bar t}\sum_{t=1}^{\bar t} \frac{3\kappa_0+1}{\sqrt{t}}\label{eq:proof-inexact-17}\\
&\leq \frac{2\kappa_0}{\sqrt{\bar t}} + \frac{1}{\bar t}\sum_{t=1}^{\bar t} \frac{3\kappa_0+1}{\sqrt{t}}\label{eq:proof-inexact-18}\\
&\leq \frac{6\kappa_0}{\sqrt{\bar t}}.\label{eq:proof-inexact-19}
\end{align}
Here, Eq.~(\ref{eq:proof-inexact-16}) holds simply by the definition of $\{\bar r_t\}_t$;
Eq.~(\ref{eq:proof-inexact-17}) holds by invoking Eq.~(\ref{eq:proof-inexact-15-half}); 
Eq.~(\ref{eq:proof-inexact-18}) holds by incorporating the $\mathcal B,\mathcal C,\mathcal D$ terms from Eq.~(\ref{eq:proof-inexact-7}) and Eqs.~(\ref{eq:proof-inexact-11},\ref{eq:proof-inexact-13},\ref{eq:proof-inexact-14}).
Eq.~(\ref{eq:proof-inexact-19}) and the definition of $\rho(\cdot)$ then imply the second property of Definition \ref{defn:weakly-nonstationary}. $\square$
\end{proof}

\subsection{Dynamic regret upper bound}

With Lemma \ref{lem:inexact-prox} in the previous section establishing the performance of Algorithm \ref{alg:inexact-prox} for weakly non-stationary environments, 
the following regret upper bound can be immediately established by Lemma \ref{lem:multi-scale}.

\begin{theorem}
Suppose Algorithm \ref{alg:sampling} is executed with the weakly non-stationary algorithm specified in Algorithm \ref{alg:inexact-prox} and $\rho(t)=6\kappa_0/\sqrt{t}$,
with $\kappa_0$ and other algorithm parameters being specified in Lemma \ref{lem:inexact-prox}.
Then the dynamic regret of the algorithm can be upper bounded by
$$
\mathbb E\left[\sum_{t=1}^T f_t(x_t^*)-f_t(x_t)\right]\leq \tilde O(d^2V_T^{1/3}T^{2/3} + d^{3/2}\sqrt{T}),
$$
where in the $\tilde O(\cdot)$ notation we drop polynomial dependency on other problem parameters ($\sigma,L,B_X,c_0$) and all poly-logarithmic factors.
Such polynomial dependency could be tracked by following the functions of $C_1,C_2,c_1,c_2$ in the statement of Lemma \ref{lem:multi-scale},
as well as in Lemma \ref{lem:inexact-prox} where such functions are implicitly defined.
\label{thm:dynamic-regret}
\end{theorem}

Theorem \ref{thm:dynamic-regret} shows that it is possible to achieve the $\tilde O(V_T^{1/3}T^{2/3}+\sqrt{T})$ tight dynamic regret upper bound in \cite{besbes2015non}
even when the function changing budget $V_T$ is not known a priori, thus resolving an important question in budget-adaptive non-stationary stochastic optimization.

\section{Connection between stationary and dynamic regret}\label{sec:connection}

In this section we expand our results to more general algorithms that achieve small cumulative regret comparing against \emph{stationary} benchmarks.
More specifically, for any policy $\pi$ and sequence of functions $f_1,\cdots,f_T$ that are potentially chosen adversarially, the stationary regret is defined as
\begin{equation}
\text{(stationary regret)}\;\;\;\;\;\mathbb E\left[\max_{x^*\in\mathcal X}\sum_{t=1}^T f_t(x^*)-f_t(x_t)\right].
\label{eq:defn-stationary}
\end{equation}
Comparing the definition of stationary regret with that of dynamic regret (\ref{eq:defn-dreg}), we observe that the stationary regret is always smaller than the dynamic regret
as the benchmark $x^*$ in Eq.~(\ref{eq:defn-stationary}) is common among all $T$ time periods and therefore cannot adapt to all different functions.
Such a weaker benchmark allows strong regret guarantees even when the function changes are completely uncontrolled, and has been studied extensively in the literature
as a performance measure for adversarial bandits \citep{bubeck2021kernel,auer2002nonstochastic,beygelzimer2011contextual}.

In this section, we show that any algorithm achieving satisfactory stationary regret upper bounds \emph{with high probability}
 can be automatically converted to an algorithm that satisfies Definition \ref{defn:weakly-nonstationary},
and would therefore enjoy similar guarantees with dynamic regret measures without knowing the changing budget a priori.
As many existing algorithms enjoy such stationary regret guarantees, the consequence of this result could be far-reaching since it automatically converts such algorithms
(with the help of the multi-scale sampling framework in Algorithm \ref{alg:sampling}) to ones that have low dynamic regret.

More specifically, we establish the following result:
\begin{lemma}
Let $\tilde\rho(\cdot)$ be a non-increasing function such that $\tilde C(t)=t\tilde\rho(t)$ is non-decreasing.
Let $\pi$ be a policy that satisfies, for every $t\geq 1$, with probability $1-\tilde O(T^{-2})$ that
\begin{equation}
\max_{x^*\in\mathcal X}\sum_{\tau=1}^t f_\tau(x^*)-f_\tau(x_\tau)\leq \tilde\rho(t).
\label{eq:stationary-regret}
\end{equation}
Then the statistics $\bar r_t := \frac{1}{t}\sum_{\tau=1}^t y_\tau + \tilde\rho(t) + \sqrt{\frac{\ln(2T)}{t}}$ satisfy Definition \ref{defn:weakly-nonstationary}
with $\rho(t)=2\tilde\rho(t)+3\sqrt{\ln(2T)/T}$ and $\lambda=2$.
\label{lem:converting}
\end{lemma}

Lemma \ref{lem:converting} shows that any algorithm achieving a cumulative regret upper bound of $\tilde\rho(t)$ comparing against stationary benchmarks
can be easily converted into a UCB algorithm in weakly non-stationary environments with a slightly inflated regret measure $\rho(t)$,
which with the help of the multi-scale sampling algorithm \ref{alg:sampling} yields a dynamic regret upper bound of $\tilde O(\mathrm{poly}(d)\times (V_T^{1/3}T^{2/3}+\sqrt{T}))$ if $\tilde\rho(t)\asymp 1/\sqrt{t}$.
This result can be immediately applied to ellipsoidal methods such as those developed in \citep{bubeck2021kernel},
attaining a dynamic regret of $\tilde O(\mathrm{poly}(d)\times (V_T^{1/3}T^{2/3}+\sqrt{T}))$ without strong smoothness or concavity assumptions on individual objective functions.
This result is also directly applicable to the EXP4 algorithm for linear bandit \citep{auer2002nonstochastic} provided that the action spaces do not change.

In the rest of this section we give the complete proof of Lemma \ref{lem:converting}.

\begin{proof}{Proof of Lemma \ref{lem:converting}.}
Fix an arbitrary $t\in[T]$. By Hoeffding-Azuma inequaltiy \citep{hoeffding1994probability,azuma1967weighted}, it holds with probability $1-\tilde O(T^{-2})$ that
\begin{equation}
\left|\frac{1}{t}\sum_{\tau=1}^t y_\tau - f_\tau(x_\tau)\right|\leq \sqrt{\frac{\ln(2T)}{t}}.
\label{eq:proof-converting-1}
\end{equation}
Let $\mu=\arg\min_{\mu\leq t}f_\mu^*$. Then with probability $1-\tilde O(T^{-2})$, we have
\begin{align}
\bar r_t &\geq \frac{1}{t}\sum_{\tau=1}^t f_\tau(x_\tau) + \tilde\rho(t) \geq \frac{1}{t}\sum_{\tau=1}^t f_\tau(x_\mu^*) \label{eq:proof-converting-2}\\
&\geq \frac{1}{t}\sum_{\tau=1}^t f_\mu(x_\mu^*) - \Delta_{[1:t]} = f_\mu^* - \Delta_{[1:t]},
\end{align}
which proves the first property in Definition \ref{defn:weakly-nonstationary}.
Here, Eq.~(\ref{eq:proof-converting-2}) holds by applying Eqs.~(\ref{eq:stationary-regret},\ref{eq:proof-converting-1}).

We next prove the second property. Note that for each $\tau$, 
\begin{align}
\bar r_\tau &\leq \frac{1}{\tau}\sum_{j=1}^t f_j(x_j) + \tilde\rho(\tau) + 2\sqrt{\frac{\ln(2T)}{\tau}}\label{eq:proof-converting-3}\\
&\leq \frac{1}{\tau}\sum_{j=1}^\tau f_j(x_j^*) + \tilde\rho(\tau) + 2\sqrt{\frac{\ln(2T)}{\tau}}\nonumber\\
&\leq \frac{1}{\tau}\sum_{j=1}^\tau f_1(x_j^*) + \Delta_{[1,\tau]} + \tilde\rho(\tau) + 2\sqrt{\frac{\ln(2T)}{\tau}}\nonumber\\
&\leq \frac{1}{\tau}\sum_{j=1}^\tau f_1(x_1^*) + \Delta_{[1,\tau]} + \tilde\rho(\tau) + 2\sqrt{\frac{\ln(2T)}{\tau}}\nonumber\\
&= \frac{1}{t}\sum_{\tau=1}^t f_1(x_1^*) + \Delta_{[1,\tau]} + \tilde\rho(\tau) + 2\sqrt{\frac{\ln(2T)}{\tau}}\nonumber\\
&\leq \frac{1}{t}\sum_{\tau=1}^t f_\tau(x_1^*) + \Delta_{[1,\tau]} + \Delta_{[1,t]} + \tilde\rho(\tau) + 2\sqrt{\frac{\ln(2T)}{\tau}}\nonumber\\
&\leq \frac{1}{t}\sum_{\tau=1}^t f_\tau(x_\tau)+ \Delta_{[1,\tau]} + \Delta_{[1,t]} + 2\tilde\rho(\tau) + 2\sqrt{\frac{\ln(2T)}{\tau}}\label{eq:proof-converting-4}\\
&\leq \frac{1}{t}\sum_{\tau=1}^t y_\tau + \Delta_{[1,\tau]} + \Delta_{[1,t]} + 2\tilde\rho(\tau) + 3\sqrt{\frac{\ln(2T)}{\tau}}.\label{eq:proof-converting-5}
\end{align}
Here, in Eqs.~(\ref{eq:proof-converting-3},\ref{eq:proof-converting-5}) we apply Eq.~(\ref{eq:proof-converting-1});
in Eq.~(\ref{eq:proof-converting-4}) we apply Eq.~(\ref{eq:stationary-regret}).
Subsequently, noting that $\Delta_{[1,\tau]}\leq \Delta_{[1,t]}$ for all $\tau\leq t$, summing both sides of Eq.~(\ref{eq:proof-converting-5}) over $\tau=1,2,\cdots,t$ and re-arranging terms we obtain
$$
\frac{1}{t}\sum_{\tau=1}^t(\bar r_\tau-y_\tau) \leq 2\tilde\rho(\tau) + 3\sqrt{\frac{\ln(2T)}{\tau}} + 2\Delta_{[1,t]},
$$
which proves the second property of Definition \ref{defn:weakly-nonstationary} with $\lambda=2$ and the definition of $\rho(\cdot)$. $\square$
\end{proof}

\section{Concluding remarks}

In this paper we study the non-stationary stochastic optimization problem with bandit (zeroth-order) feedback, when the function change budget $V_T$ is unknown a priori.
By designing a fixed-step-size estimating gradient descent algorithm and coupling it with the multi-scale sampling framework developed in \citep{wei2021non}, we achieve a dynamic regret upper bound
of $\tilde O(V_T^{1/3}T^{2/3}+\sqrt{T})$ without knowing $V_T$, which matches the information-theoretic lower bound established in \citep{besbes2015non}.
We also derive additional result connecting stationary regret with dynamic regret, which is applicable to an even wider range of bandit problems and algorithms \citep{bubeck2021kernel,auer2002nonstochastic}.

To conclude this paper, we mention a question that invites future research. 
It is known that the standard estimating gradient descent algorithm with $\eta_t\propto 1/t$ diminishing step sizes achieves $\tilde O(\sqrt{T})$ regret comparing against stationary benchmarks,
even when the underlying function sequence is completely arbitrary (see, e.g.~the analysis in \citep{flaxman2005online,agarwal2010optimal}).
However, since the algorithm only enjoys good stationary regret bounds in expectation, it cannot be directly used to satisfy Definition \ref{defn:weakly-nonstationary}.
On the other hand, the estimating gradient descent method proposed in this paper using fixed step sizes and geometrically increasing epoch sizes works well only in weakly non-stationary environments,
and it is easy to see that the method is not effective when the underlying function changes arbitrarily (in which case the gradient estimates $\hat g_s$ in Algorithm \ref{alg:inexact-prox}) degrade arbitrarily.
Hence, the methods developed in this paper achieve $\tilde O(V_T^{1/3}T^{2/3}+\sqrt{T})$ dynamic regret without knowing $V_T$, but it is unlikely that they achieve $\tilde O(\sqrt{T})$ stationary regret
in completely adversarial environments. It is thus an interesting future research question to design ``best-of-both-worlds'' algorithm that simultaneously achieves $\tilde O(\sqrt{T})$ stationary regret
in fully adversarial settings, and $\tilde O(V_T^{1/3}T^{2/3}+\sqrt{T})$ dynamic regret when function changes are controlled by $V_T$.
It is likely that certain interpolation between estimating gradient descent algorithms with fixed and diminishing step sizes is required to achieve such ``best-of-both-world'' results.

\bibliographystyle{ormsv080}
\bibliography{refs}

\end{document}